\documentclass[sigconf]{acmart}
\AtBeginDocument{%
 \providecommand\BibTeX{{%
  \normalfont B\kern-0.5em{\scshape i\kern-0.25em b}\kern-0.8em\TeX}}}

\usepackage{tabularx}
\usepackage{multicol} 
\usepackage{multirow}
\usepackage{threeparttable}
\usepackage{float}
\restylefloat{table}
\usepackage{amsthm}
\usepackage{amsmath,amsfonts,bm}

\makeatletter
\def\BState{\State\hskip-\ALG@thistlm}
\makeatother

\def\eqref#1{equation~\ref{#1}}
\def\1{\mathbb{I}}

\def\rz{{\textnormal{z}}}

\def\ry{{\textnormal{y}}}

\def\rvz{{\mathbf{z}}}

\def\vf{{\bm{f}}}

\def\vh{{\bm{h}}}

\def\vp{{\bm{p}}}

\def\vz{{\bm{z}}}

\def\vx{{\bm{x}}}

\DeclareMathAlphabet{\mathsfit}{\encodingdefault}{\sfdefault}{m}{sl}
\SetMathAlphabet{\mathsfit}{bold}{\encodingdefault}{\sfdefault}{bx}{n}

\newcommand{\R}{\mathbb{R}}
\newcommand{\N}{\mathbb{N}}

\begin{document}

%%
%% The "title" command has an optional parameter,
%% allowing the author to define a "short title" to be used in page headers.
\title{Predicting Legal Proceedings Status: Approaches Based on Sequential Text Data}

%%
%% The "author" command and its associated commands are used to define
%% the authors and their affiliations.
%% Of note is the shared affiliation of the first two authors, and the
%% "authornote" and "authornotemark" commands
%% used to denote shared contribution to the research.

\author{Felipe Maia Polo}
\email{felipemaiapolo@gmail.com}
\affiliation{%
 \institution{University of São Paulo \\ Advanced Institute for AI (AI2)}
 \city{São Paulo}
 \state{São Paulo}
 \country{Brazil}
}

\author{Itamar Ciochetti}
\email{itamar@tikal.tech}
\affiliation{%
 \institution{Tikal Tech}
 \city{São Paulo}
 \state{São Paulo}
 \country{Brazil}
}

\author{Emerson Bertolo}
\email{emerson@tikal.tech}
\affiliation{%
 \institution{Tikal Tech}
 \city{São Paulo}
 \state{São Paulo}
 \country{Brazil}
}

%%
%% The abstract is a short summary of the work to be presented in the
%% article.
\begin{abstract}
  The objective of this paper is to develop predictive models to classify Brazilian legal proceedings in three possible classes of status: (i) archived proceedings, (ii) active proceedings, and (iii) suspended proceedings. This problem's resolution is intended to assist public and private institutions in managing large portfolios of legal proceedings, providing gains in scale and efficiency. In this paper, legal proceedings are made up of sequences of short texts called "motions." We combined several natural language processing (NLP) and machine learning techniques to solve the problem. Although working with Portuguese NLP, which can be challenging due to lack of resources, our approaches performed remarkably well in the classification task, achieving maximum accuracy of $.93$ and top average F1 Scores of $.89$ (macro) and $.93$ (weighted). Furthermore, we could extract and interpret the patterns learned by one of our models besides quantifying how those patterns relate to the classification task. The interpretability step is important among machine learning legal applications and gives us an exciting insight into how black-box models make decisions.
\end{abstract}

\keywords{Law, Machine Learning, Natural Language Processing, Text Sequences}

%% This command processes the author and affiliation and title
%% information and builds the first part of the formatted document.
\maketitle

%%%% Format %%%%
% 1. Intro
% 2. Objective and practical importance 
% 3. Related Work
% 4. Data
% 5. Methodology
%  5.1 (Generic) Neural Network architecture (texts -> features -> lstm)
%  5.2 Feature Extraction 
%    5.2.1 Default Text preprocessing
%    5.2.2 Text embedding approaches 
%    5.2.3 Token embedding approach
% 6. Classification results 
% 7. Interpretability
%  7.1 Mathematical details of the CNN architecture
%  7.2 Interpreting the filters
%  7.3 Results
% 8. Conclusion

%TO-DO
%1. revisar metricas e bootstrap (fazer resampling no teste)
% criar funções p termos menos código
%
%
%

\section{Introduction}

Machine learning (ML) and natural language processing (NLP) tools are present in many fields and can perform diverse tasks efficiently. One field that is already undergoing significant changes, and there is still much room to work in the coming years, is Law. This paper makes extensive use of ML and NLP to classify Brazilian legal proceedings, represented by texts sequences. Our efforts aim to develop classification models to predict a concrete fact in the legal field: legal proceedings status. We combine several techniques to analyze sequences of texts in chronological order, which are common in the legal context. Moreover, we explore how one of our approaches can generate interpretations and insights, which can be very useful within the legal context.

This paper is organized as follows. In Section \ref{sec:obj}, we give details about the objectives of this work in addition to explaining its relevance. In Section \ref{sec:rev}, we briefly review papers related to this work. In Section \ref{sec:data}, we present the datasets used in our experiments. In Section \ref{sec:method}, we give more details about the methodologies used for the classification of legal proceedings and, soon after, we present the results in Section \ref{sec:result}. Finally, in Section \ref{sec:interp}, we explain how to obtain interpretability from one of our approaches and present insights from that neural network model.
\section{Objective and practical importance of this work}
\label{sec:obj}

The objective of this paper is to develop predictive models for the classification of Brazilian legal proceedings in three possible classes of status: (i) archived proceedings, (ii) active proceedings, and (iii) suspended proceedings. Each proceeding is made up of a sequence of short texts called “motions” written in Portuguese by the courts’ administrative staff. The motions relate to the proceedings, but not necessarily to their legal status. The three possible classes are given in a specific instant in time, which may be temporary or permanent. Moreover, they are decided by the courts to organize their workflow, which in Brazil may reach thousands of simultaneous cases per judge. Besides building a good classifier, we also value the interpretability of the results achieved, given the importance of understanding the decisions made by models in the legal field.

The objective of classifying legal processes according to their status classes (archived, active, and suspended) was chosen due to four complementary reasons: (1) the importance of public and private institutions knowing the status of legal proceedings of their interest, (2) the universality of the problem, due to the overwhelming amount of cases in Brazil and the widespread use of this categorization for the basic organization of the workflow of clerks and secretariats, (3) the difficulty in obtaining status information directly from the courts, and (4) the long time it may take for a lawyer to classify proceedings according to their status manually. More details are given in the following:

\begin{enumerate}
  \item Institutions such as large companies, governments, law firms, and legal/law techs often have to manage many of their own or clients’ proceedings. The status is the most basic information needed to manage legal proceedings, as this information dictates the possible types of actions that these institutions may take towards their clients or Justice;
  
  \item Even though there are more than 90 different courts in Brazil (State, Labor, Federal, and others), all Brazilian legal proceedings must be classified in one of the three classes of status. This universality of status labels is probably also true in many other countries;
  
  \item Despite status being a piece of objective information, it is not easy to access in most cases. When there is a need to manage many proceedings, such as in law firm portfolios, seeking this information directly from the legal system is not feasible. There are four main reasons for that: (i) this type of information may be spread on dozens of different individual courts’ web pages, (ii) the access is not usually straightforward given the courts might have made their websites hard to scrap, (iii) the information is non-structured and non-standardized in most cases, due to the use of many variations of the description of the three status under the different courts’ jargons, and (iv) the information maybe even inaccurate, incorrect and outdated;

  \item On average, a lawyer takes around 3 minutes to classify a case according to its status. Let us consider a mass of 6,500 cases that must be labeled, our dataset’s size in this paper. A lawyer would take around 13 uninterrupted days with the labeling process, on average. On the other hand, our machine learning and NLP models take less than five minutes to do all the classification steps on an ordinary laptop while maintaining an accuracy of approximately $93\%$.
\end{enumerate}

In summary, the design of a classifier that predicts a legal proceeding’s status can be a great ally in gaining efficiency and scale in decentralized and non-standardized legal systems such as that of Brazil and other developing countries.

\section{Related work}
\label{sec:rev}

It is a fact that efforts to apply machine learning and natural language processing techniques to solve legal problems are increasing. Most of the applications are made to predict or understand legal results. Still, some applications aim to make the legal system more efficient, acting directly on a more administrative front, which is this paper's primary concern. Some works in the literature aim to have interpretable results in addition to creating predictive models. That is also our case.

Despite ML researchers' efforts to create applications in the legal field, we were unable to find an attempt to solve a problem like ours in the literature. The issues closest to ours we could relate in literature are those of identifying the parties in legal proceedings \cite{nguyen2018recurrent}, classification of legal documents according to their administrative labels \cite{dadocument,braz2018document} or predicting the area a proceeding belongs to \cite{sulea2017exploring}. This paper has a different application that can be useful when looking for efficiency in legal systems, especially in developing countries. Unlike previous work, we consider sequences of texts explicitly in our modeling, which has not yet been observed in Law and AI literature by us.

Because Law is directly linked to high stakes decisions, a significant concern is to create interpretable or explainable models. In this sense, some recent works have been developed as applications in the legal field \cite{westermann2019using, marques2019machine, chalkidis2018obligation}. On the other hand, these works (i) require a high level of feature engineering, (ii) do not provide a big picture interpretation, in contrast to explaining particular data points, or (iii) are not directly adaptable to sequences of texts. This paper contributes to the literature as it uses almost no feature engineering and applies simple concepts and tools such as cosine similarity and partial dependence plots for an intuitive interpretation of general results in the classification of text sequences, which can be applied beyond the legal field.
\section{Data}
\label{sec:data}

Our data is composed of two datasets: a dataset of $ 3 \cdot 10^6 $ unlabeled motions and a dataset containing $ 6449 $ legal proceedings, each with an individual and a variable number of motions, but which have been labeled by lawyers. Among the labeled data, $47.14\%$ is classified as archived (class 1), $45.23\%$ is classified as active (class 2), and $7.63\%$ is classified as suspended (class 3). 

In order to make things more concrete, we should mention that the motions have a specific format; an example is \textit{"Type of Motion: Ordinary Act Practiced Description: Be aware of the Court's record. Wait for the interested party's manifestation. Nothing being requested, the records will be forwarded to DIPEA."}

The datasets we use are representative samples from the first (São Paulo) and third (Rio de Janeiro) most significant state courts. State courts handle the most variable types of cases throughout Brazil and are responsible for $80\%$ of the total amount of lawsuits. Therefore, these datasets are a good representation of a very significant portion of the use of language and expressions in Brazilian legal vocabulary. 

Since classifying sequences of texts is a complex task and our dataset of labeled proceedings is not very large, we pretrain token and text embeddings for the legal domain with the unlabeled dataset. Then, we use the labeled dataset to create predictive models for the legal proceedings classification. 

\section{Methodology}
\label{sec:method}

We used four approaches to extract features from the legal texts and three base classifiers to create our predictive models to classify legal proceedings.

In this section, we address the content in the following order: (i) the classifiers architecture, (ii) the four ways we use to extract features from texts, (iii) the dataset split, and (iv) the hyperparameters tuning phase.

\subsection{Classifiers}

In Section \ref{sec:obj}, we mentioned that each of the legal proceedings comprises a chronological sequence of short texts called motions. Assuming that each text is represented by a vector of features and there is a temporal structure among texts that must be respected, it is natural to think that some architecture involving many-to-one recurrent neural networks (RNNs) is a reasonable solution. Within the RNN literature, long short-term memory neural networks (LSTM) \cite{hochreiter1997long} are perhaps the best known and most successful architectures for solving a series of problems present in classic RNNs, such as vanishing gradients and long-term memory. For this reason, we use the many-to-one LSTM architecture as a classifier. The inputs are given by vectors representing texts, and the outputs will be predicted probabilities for each of the three classes, returned by the Softmax function. 

It is possible to see in Figure \ref{fig:nn_arc} a diagram with the architecture used, where "Text -1" is the most recent motion and "Text -T" is the least recent one to be considered:

\begin{figure}[h]
 \centering
 \includegraphics[width=.35\textwidth]{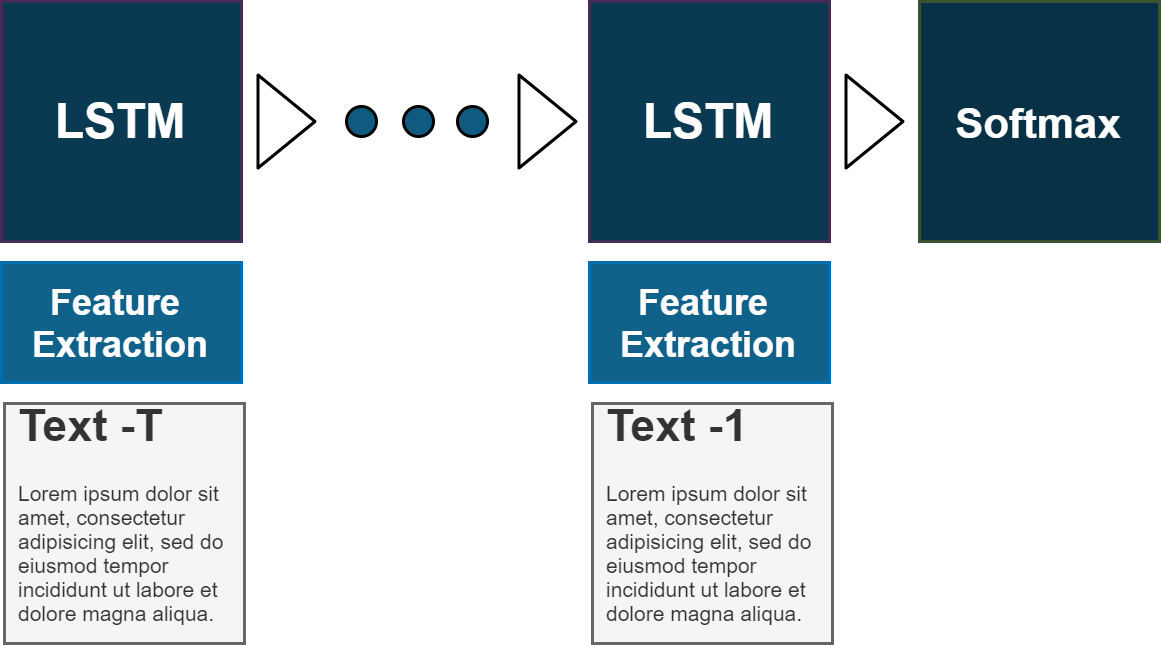}
 \caption{First classifier architecture: to create our predictive models for legal proceedings classification, we used four approaches to extract features from legal texts and the above base architecture as a classifier.}
 \label{fig:nn_arc}
\end{figure}

We adopted two other possible classifiers in addition to our first LSTM neural network. A multilayer perceptron neural network (MLP) \cite{goodfellow2016deep} with one hidden layer and ReLU activation functions gives the first one. Our other classifier is given by an XGBoost \cite{chen2016xgboost} tree ensemble, state of the art in classification and regression tasks with tabular data. As these last two classification models do not allow, a priori, a temporal structure for the data, we concatenate the feature vectors of the last $ T $ texts to feed the models.

Figure \ref{fig:arc} shows a diagram with the architecture used, where "Text -1" is the most recent motion and "Text -T" is the least recent one to be considered:

\begin{figure}[h]
 \centering
 \includegraphics[width=.25\textwidth]{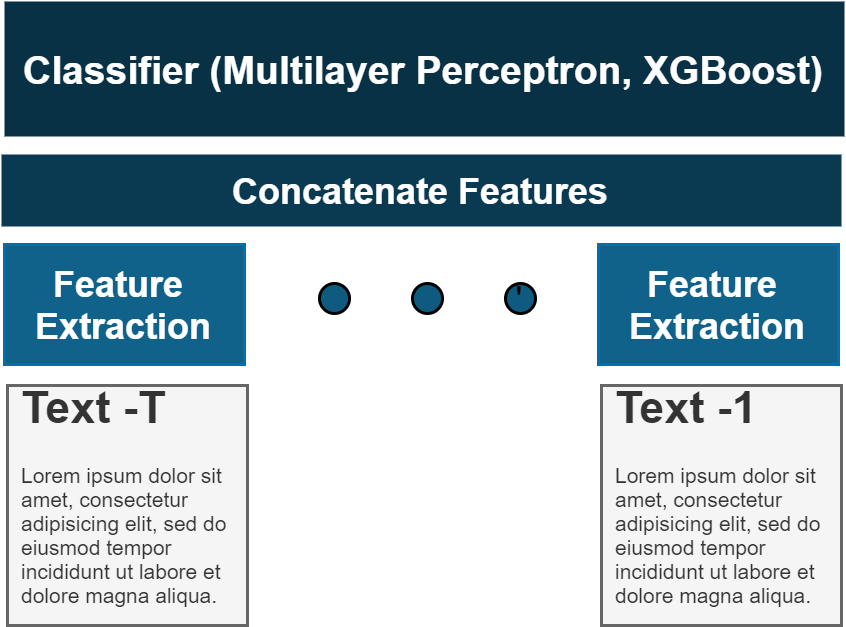}
 \caption{Second/third classifier architecture: to create our predictive models for legal proceedings classification, we used four approaches to extract features from legal texts and the above base architecture as a classifier.}
 \label{fig:arc}
\end{figure}

All classifiers make classification choosing the most probable class.

The only remaining point in this section is to decide the number of texts that we use as input in our classifiers. Our experience working with motions tells us that keeping only the last five texts is sufficient for the task. In some tests, we verify that the models did not improve their performance by adding more than five texts, which confirmed our intuition. This fact guarantees that working with the last five motions in each legal proceedings would give us parsimonious and satisfying solutions, so we fix $T=5$. For cases where the proceedings have less than five texts, we use zero-padding vectors to complete the inputs.

\subsection{Feature Extraction}

We use four different approaches to extract features from texts, all of them being completely unsupervised or self-supervised (Word2Vec (W2V) \cite{mikolov2013efficient}, Doc2Vec (D2V) \cite{le2014distributed}, TFIDF \cite{salton1986introduction} and BERT \cite{devlin2018bert}). For all approaches, we had a preprocessing step for the texts. In the following, we present the procedures for cleaning and tokenizing the texts and give more details about the four different approaches to extract features from the texts.

\subsubsection{Text Preprocessing}\label{sec:prep}

Before applying any natural language processing or machine learning model to text, a preliminary step is preprocessing the raw data obtained in text form. We applied the four points below, which are standard in the literature of NLP:

\begin{enumerate}
 \item \textit{Uppercase to lowercase conversion (Word2Vec, Doc2Vec, TFIDF)}: we standardize the words in the body of the texts by converting uppercase characters to lowercase;
 
 \item \textit{Undesired symbols removal (Word2Vec, Doc2Vec, TFIDF)}: we remove all punctuation and graphic accentuation (a very few were also removed when using BERT);
 
 \item \textit{Tokenization (All)}: 
 \begin{itemize}
  \item For BERT approach: we used the same tokenizer used by \citet{souza2019portuguese};
  \item For Word2Vec, Doc2Vec, TFIDF approaches: we use the method proposed by \citet{mikolov2013distributed} in order to identify presence words that generally appear together and which should be considered as unique tokens\footnote{This method is implemented in the Gensim package \url{https://radimrehurek.com/gensim/}.}. Applying this method twice in sequence, with default threshold, we could identify which sets of 2 to 4 words should be considered as unique tokens.
 \end{itemize}
 
 \item \textit{Standardization of expressions (All)}: Examples are the conversion of the terms "state law" and "federal law" to "law";
\end{enumerate}

\subsubsection{Word2Vec approach}

In the following, we detail Word2Vec training for tokens embeddings and then explain how we obtain representations for texts using those embeddings. The construction of tokens embeddings in this work is entirely self-supervised using a mass of $ 3 \cdot 10^6 $ texts/motions from unlabeled proceedings. We made that choice since we have only a small number of labeled text sequences. Once we have the mass of preprocessed texts, we use the model specified in \cite{mikolov2013efficient} (Continuous Bag of Words Word2Vec) (size $ D $ = 100, window = 10) \footnote{We tested many configurations, e.g., windows = 5, 10, 15 and size = 50, 100, 150, and we chose to work with the more parsimonious and best performing one, according to the classification results.} and extract the vector representations for each of the tokens in the vocabulary. After obtaining each of the vector representations of tokens, we normalize them to have a unitary euclidean norm, facilitating interpretability, as we show in Section \ref{sec:interp}.

To create the representations for the texts, we use two different approaches: one of them is used in conjunction with the LSTM classifier, and the other is used in conjunction with the MLP and XGBoost classifiers. In both approaches, we use the first $ R = 70 $ tokens\footnote{We have noticed that $ 95 \% $ of the motions have a maximum of 70 tokens, and the critical information is not likely to be at the end of the texts.} for each of the texts, completing texts using null vectors of size $ D =100$ when needed. 

When using LSTM networks as classifiers, we first set up a matrix with dimensions $ R \times D $ for each of the texts, each row being given by a token embedding. Secondly, we apply $ K $ one-dimensional convolutional filters \cite{kim2014convolutional} to extract the desirable information from the texts. The filters are trained in conjunction with the LSTM weights. Due to the last detail, we also refer to the W2V/LSTM classification approach as W2V/CNN/LSTM, where the "CNN" stands for convolutional neural networks. In the neural network's learning process, we constrain the euclidean norm of filters equal to one and freeze the embeddings. We give a more detailed explanation on this architecture in Section \ref{sec:interp}.

On the other hand, to represent texts when working with MLP and XGBoost classifiers, we use the average vector of each text's tokens embeddings. In other words, we collect all the non-null vectors that represent tokens in a text and calculate their average. The average vectors of all last five motions are then concatenated to represent a proceeding.

\subsubsection{Doc2Vec, TFIDF, and BERT approaches}

The other three ways to extract features are applications of the Doc2Vec \cite{le2014distributed}, TFIDF \cite{salton1986introduction} and BERT-Base\footnote{Feature-based approach with 768 features.} \cite{devlin2018bert} models. For the Doc2Vec alternative, we kept the specifications for the Word2Vec model presented in the last section. Actually, both Word2Vec and Doc2Vec models were trained together. For the TFIDF alternative, we trained the representation using the unlabeled dataset, imposing a ceiling of 4000 tokens. That is, keeping the most 4000 frequents tokens in the corpus. Regarding the BERT alternative, we fine-tuned the Portuguese BERT-Base model pre-trained by \citet{souza2019portuguese} using the Masked Language Model (MLM) objective on the unlabelled dataset, keeping the same model configuration and vocabulary used by the authors. We trained the model by adopting one epoch, batch size of 4 texts, MLM probability of 0.15, and a Tesla T4 GPU. The optimizer configuration was kept to Hugging Face's Transformer default\footnote{As in the file \url{https://github.com/huggingface/transformers/blob/master/src/transformers/training_args.py}}.

\subsection{Training, validation and test sets}

In order to train and assess our classifiers, we split at random our labeled dataset into three parts: training set ($70\%$), validation set ($10\%$), and test set ($20\%$). We used the training set to fit the model, the validation set to choose the best hyperparameters, and the test set to check the final model's performance. The final models are trained in the training set only once (after the hyperparameter tuning phase) while we compute the test scores using the test set and then bootstrapping it 100 times to get the scores standard errors.

\subsection{Hyperparameter tuning}

Hyperparameters are parameters used to control models and algorithms behavior and are not learned by the algorithms themselves \cite{goodfellow2016deep}. We have chosen to keep some of the hyperparameters fixed and tune the rest of them in a simple holdout validation procedure using the random search approach. We train and validate our models using $50$ combinations of values chosen randomly from a grid. Then we select that combination that returns the highest accuracy.

Firstly we focus on the neural networks classifier, i.e., LSTM and MLP. We fixed the following hyperparameters for both architectures: optimizer Adam (Keras default configuration and learning rate equals .005), 50 epochs, and batch size 500. The other most important hyperparameters are validated according to the values in Table \ref{hyper1}. A remark is that the number of convolutional filters only matters for the W2V/CNN/LSTM approach.
\begin{table}[h]
 \centering
 \caption{Possible Values hyperparameters for the classifiers LSTM and MLP. The number of convolutional filters only matters for our W2V/CNN/LSTM}
 \medskip
 \begin{tabular}{c|c}
 \toprule
 Hyperparameter & Possible Values \\
 \midrule 
 \# Convolutional filters (K) & $3,~6,~9$ \\[.25em]
 Hidden state size (H) & $10,~25,~50,~75,~100,~150,~200$ \\[.25em]
 LSTM/MLP weights l1 & $0$, $10^{-6}$, $5\cdot 10^{-6}$, $10^{-5}$,  \\
 and l2 penalization strength &$5\cdot10^{-5}$, $10^{-4}$, \\
 (Elastic Net regularization)&$5\cdot10^{-4}$, $10^{-3}$, $5\cdot10^{-3}$ \\
 \bottomrule
 \end{tabular}%
 \label{hyper1}%
\end{table}%

For the XGBoost classifier, we fix the use of 500 boosted trees and early stopping rounds of 15. The other most important hyperparameters were validated according to the values in Table \ref{hyper2}. The hyperparameters' names are reported as they are in the XGBoost Python API\footnote{See \url{https://xgboost.readthedocs.io/en/latest/python/python_api.html}}.

\begin{table}[h]
 \centering
 \caption{Possible Values hyperparameters for the XGBoost classifiers}
 \medskip
 \begin{tabular}{c|c}
 \toprule
 Hyperparameter & Possible Values \\
 \midrule 
 Max. depth & $3,~5,~7,~9$ \\[.25em]
 Learning rate & $.1,~.5,~1.$ \\[.25em]
 Gamma & $0, ~10^{-5}, ~10^{-4}, ~10^{-3}, ~10^{-2}, ~10^{-1}, ~1$  \\[.25em]
 Lambda & $10., ~50., ~100., ~300., ~500.,~ 1000.$  \\
 \bottomrule
 \end{tabular}%
 \label{hyper2}%
\end{table}%

The best values for hyperparameters can be found in the appendix.

\section{Classification Results}
\label{sec:result}

\begin{table*}[ht] 
 \centering 
 \caption{Evaluation of classification approaches (scores $\pm$ bootstrap std. errors). We combine three basic classifiers (LSTM, MLP, and XGBoost) and four approaches for extracting features (Word2Vec, Doc2Vec, TFIDF, and BERT). We achieved satisfying results in all our approaches, except for the \textit{Doc2Vec} ones, that performed relatively worse than the others.} 
  \begin{tabular}{c|c|c|ccc|ccc|c} 
  \hline 
 \multicolumn{3}{c}{} & \multicolumn{3}{c}{Macro averaging} & \multicolumn{3}{c}{Weighted averaging} & \multicolumn{1}{c}{} \\ 
 \hline 
  Classifier & Feature extraction & Accuracy & F1 Score & Precision & Recall & F1 Score & Precision & Recall \\ 
  \hline 
 \multirow{4}{*}{LSTM}   & W2V (using CNN) &$\mathbf{0.93  \pm  0.01}$&0.88 $ \pm $ 0.01&0.92 $ \pm $ 0.01&0.85 $ \pm $ 0.02&0.92 $ \pm $ 0.01&$\mathbf{0.93  \pm  0.01}$&$\mathbf{0.93  \pm  0.01}$\\ 
  & Doc2Vec &0.82 $ \pm $ 0.01&0.76 $ \pm $ 0.02&0.77 $ \pm $ 0.02&0.75 $ \pm $ 0.02&0.82 $ \pm $ 0.01&0.82 $ \pm $ 0.01&0.82 $ \pm $ 0.01\\ 
  & TFIDF &0.90 $ \pm $ 0.01&0.85 $ \pm $ 0.01&0.85 $ \pm $ 0.01&0.85 $ \pm $ 0.02&0.90 $ \pm $ 0.01&0.90 $ \pm $ 0.01&0.90 $ \pm $ 0.01\\ 
  & BERT &$\mathbf{0.93  \pm  0.01}$ &$\mathbf{0.89  \pm  0.01}$&0.92 $ \pm $ 0.01&$\mathbf{0.87  \pm  0.02}$&$\mathbf{0.93  \pm  0.01}$&$\mathbf{0.93  \pm  0.01}$&$\mathbf{0.93  \pm  0.01}$\\ 
  \hline 
 \multirow{4}{*}{MLP}    & W2V &0.91 $ \pm $ 0.01&0.87 $ \pm $ 0.01&0.92 $ \pm $ 0.01&0.84 $ \pm $ 0.02&0.91 $ \pm $ 0.01&0.91 $ \pm $ 0.01&0.91 $ \pm $ 0.01\\ 
  & Doc2Vec &0.81 $ \pm $ 0.01&0.76 $ \pm $ 0.02&0.78 $ \pm $ 0.02&0.74 $ \pm $ 0.02&0.81 $ \pm $ 0.01&0.81 $ \pm $ 0.01&0.81 $ \pm $ 0.01\\ 
  & TFIDF &0.92 $ \pm $ 0.01&0.87 $ \pm $ 0.01&$\mathbf{0.93  \pm  0.01}$&0.84 $ \pm $ 0.02&0.92 $ \pm $ 0.01&0.92 $ \pm $ 0.01&0.92 $ \pm $ 0.01\\ 
  & BERT &0.89 $ \pm $ 0.01&0.83 $ \pm $ 0.02&0.91 $ \pm $ 0.01&0.79 $ \pm $ 0.02&0.89 $ \pm $ 0.01&0.89 $ \pm $ 0.01&0.89 $ \pm $ 0.01\\ 
  \hline 
 \multirow{4}{*}{XGBoost}    & W2V &0.92 $ \pm $ 0.01&0.87 $ \pm $ 0.01&0.92 $ \pm $ 0.01&0.84 $ \pm $ 0.02&0.92 $ \pm $ 0.01&0.92 $ \pm $ 0.01&0.92 $ \pm $ 0.01\\ 
  & Doc2Vec &0.87 $ \pm $ 0.01&0.83 $ \pm $ 0.01&0.89 $ \pm $ 0.01&0.79 $ \pm $ 0.02&0.87 $ \pm $ 0.01&0.88 $ \pm $ 0.01&0.87 $ \pm $ 0.01\\ 
  & TFIDF &0.92 $ \pm $ 0.01&0.88 $ \pm $ 0.01&$\mathbf{0.93  \pm  0.01}$&0.84 $ \pm $ 0.02&0.92 $ \pm $ 0.01&$\mathbf{0.93  \pm  0.01}$&0.92 $ \pm $ 0.01\\ 
  & BERT &0.92 $ \pm $ 0.01&0.86 $ \pm $ 0.01&0.92 $ \pm $ 0.01&0.83 $ \pm $ 0.02&0.92 $ \pm $ 0.01&0.92 $ \pm $ 0.01&0.92 $ \pm $ 0.01\\ 
 \hline 
  \end{tabular}% 
 \label{tab:metrics}% 
 \end{table*}% 
 
Given that all classifiers make classification choosing the most probable class, we now compare their performance according to a few key metrics. The final models are trained in the training set only once (after the hyperparameter tuning phase) while we compute the test scores using the test set and then bootstrapping it 100 times to get the scores standard errors.

The first metric we use to evaluate and compare each of our classifiers' performance is accuracy. Accuracy gives us the percentage of correct predictions of a classifier. Thus, it is a metric capable of giving us a general idea of how the models perform. In the third column of Table \ref{tab:metrics} one can see how our 12 approaches, combining three different classifiers and four ways to extract features, perform. If we consider a classifier that always predicts the most probable class (considering prior probabilities) as a benchmark (accuracy of .47), we achieved excellent results in all our approaches. In this case, only the \textit{Doc2Vec} approaches performed relatively worse than the others.

Although accuracy is widely used and straightforward to understand, it can generate "over-optimistic" interpretations, especially when we have a dataset with very imbalanced classes. We show in the other columns of Table \ref{tab:metrics} averages for the Precision, Recall and F1 scores in order to present more robust results. We present the averages because each of these three metrics is class-specific, so the averages give us a general idea of how the classifier performs. The macro averages were calculated merely summing up the metric across classes and dividing by three. On the other hand, the weighted averages were calculated, giving different weights to the classes proportional to the number of examples in each class.

In general, our classifiers have an excellent performance. A pertinent point that can be observed in the "Recall" and "F1 Score" columns of the macro averages is that our classifiers perform relatively worse when finding examples from the "Suspended" minority class, which add up to about $ 7.63 \% $. This fact occurs precisely because this class has less weight in the loss functions when training the classifiers; however, this is not a problem as we consider that each of the samples must have the same importance as the others. If we wanted to provide more weight to the minority class elements, we could directly adjust the loss function.
 
Although the \textit{W2V}, \textit{TFIDF}, and \textit{BERT} approaches have arrived at classifiers that perform similarly, below, we further explore the \textit{W2V} + \textit{CNN/LSTM} approach. In addition to \textit{W2V/CNN/LSTM} being a great classifier, we were able to think of intuitive ways to interpret how the network works and better understand the role of specific expressions and time in the classification of legal proceedings.

\section{Interpretability}\label{sec:interp}

This section is divided into three main parts. In the first one, we explore the mathematical aspects of the W2V/CNN/LSTM model. In the second, we use this mathematical knowledge to motivate ways to interpret the results. Finally, we present the interpretability results.

\subsection{Mathematical details of the W2V/CNN/LSTM classifier architecture}
 
Let (i) $ i $ be the index of a legal proceeding\footnote{$ i $ can represent an out of sample proceeding.}, (ii) $ t \in \left \{-5, ... ~, -1 \right \} $ an index for a text/motion of $ i $ proceeding, where $ -1 $ denotes the most current text and $ -5 $ the least current text taken into account, (iii) $ n \in [70] $ an index \footnote{Consider [$N$] = $ \left \{1, ... ~, N \right \} $, $N \in \N$.} of embedded tokens in the text $ t $ from proceeding $ i $ and (iv) $ \vf_k \in \R^{100} $ is the vector representing the k-th convolutional filter, $ k \in [K] $. Then, we define the following quantity $ z_{itnk} $, which is the feature extracted by the filter $ \vf _k $ from token $ \vx_ {itn} \in \R^{100} $, that is, $n$-th token from $t$-th motion/text from $i$-th proceeding:
\begin{align}
     z_{itnk}=\vx_{itn}  \cdot \vf_k  
\end{align}
 
Where "$\cdot$" is the scalar product of two vectors. Note that we use a linear activation function in this case and removed the constant neuron, which represents the bias. Furthermore, the final feature extracted by the $\vf_k $ filter from the $t$-th motion/text from $i$-th proceeding right after applying \textit{max-over-time pooling} \cite{collobert2011natural} procedure is given by the quantity $ z^*_{itk} $ as follows:
\begin{align}
     z^*_{itk}= \textup{max} \left \{ z_{itnk} \right \}_{n=1}^{70}
\end{align}
 
Grouping those quantities through index $ k $ in an array, we have the following vector that we use to feed our recurrent neural network with LSTM units:
\begin{align}
    \vz_{i,t}^*&=(z^*_{it1},...~,z^*_{itK})
\end{align}

The probability vector of $ i $-th legal proceeding belonging to one of the three possible classes/status, $ \vp_i \in \R^3$, is given by the transformation $ \vh $ which is a recurrent neural network (RNN/LSTM) with a time depth of 5:
\begin{align}
   \vp_i=\textbf{\textit{h}}(\vz_{i,-1}^*,...~,\vz_{i,-5}^*)
\end{align}

Where $ \vz_{i, -1}^* $ refers to the most current network input and $ \vz_{i, -5}^* $ refers to the least current input. For a class $ j \in [3] $, we can also write the individual predicted probability as $ p_{ij} = h_j (\vz_{i, -1}^*, ... ~, \vz_{ i, -5}^*) $. It is not explicit, but this time, as well as all the others not mentioned, we used non-linear activation functions (LSTM default) and included the constant neuron.

\subsection{How to interpret the classifier?}

\subsubsection{What are the filters looking for?}

In the process of feature extraction performed by the convolutional layer of the network, we have that each of the K filters go through all 70 embedded representations of tokens present in each text performing scalar products. As we discussed earlier, each of the embeddings representations and filters was constrained to have a unitary euclidean norm. That means the scalar product between the filters and embeddings representations gives us the value of the cosine of the shortest angle formed between the vectors, i.e., the cosine similarity between them. Mathematically, we have:
\begin{align}
      z_{itnk}&=\vx_{itn}  \cdot \vf_k\\
       &=  \left \| \vx_{itn} \right \|  \left \| \vf_k \right \| \textup{cos}(\theta_{itnk})  \\
     &=\textup{cos}(\theta_{itnk})
\end{align}

Where $\theta_{itnk}$ is the shortest angle formed between the vectors $\vx_{itn}$ and $\vf_k$. In the learning process, the network learns representations for filters that tend to minimize the cross-entropy loss function. By constraining the vectors to have unitary euclidean norms while learning the best weights for the convolutional layer, the network aligns\footnote{By "aligning" we mean approximating according to the cosine similarity.} the filters representations to those tokens representations that help the most in minimizing the loss function. Then, by analyzing the filters' final representations, we can have insights into the patterns that the network looks for in the texts. To better understand what those patterns are, we look at the tokens with the closest representations to the filters according to cosine similarity. 
 
\subsubsection{How do features extracted by each filter relate to classification?}
 
To interpret how each filter relates to the classification task, we use Partial Dependence Plots \footnote{See \citet{molnar2019} for a more detailed explanation.}. To explain the concept, we first introduce a new notation. If $ \ry_i $ is a random variable that denotes the class of the $ i $-th proceeding, then we can rewrite $ p_{ij} $ as follows:
\begin{align}
    p_{ij}&=\hat{\mathbb{P}}\big(\ry_i=j ~\big|~ \vz_{i,-1}^*,...~,\vz_{i,-5}^*    \big) \\
    &=\hat{\mathbb{P}}\big(\ry_i=j ~\big|~ z^*_{i,-1,1},...~,z^*_{i,-5,K}  \big)
\end{align}

In order to help us define the partial dependence function, we write $\vz_{i}^*=(\vz_{i,-1}^*,...~,\vz_{i,-5}^*)$ as the concatenation of the vectors. Moreover, when we want to talk about the features themselves, i.e. random variables/vectors, and not their instances in the $ i $ individual, we can rewrite $z^*_{itk}$ as $\rz^*_{tk}$, $\vz_{i,-1}^*$ as $\rvz_{-1}^*$ and $\vz_{i}^*$ as $\rvz^*$. Given all these notations, the partial dependence function on $ \rz^*_{tk}$ feature predicting $ j $ class probability, with $ t = -1 $ and $ k = 1 $, for example, is given by:
\begin{align}
    g_{j,\rz^*_{-1,1}}(z)=\mathbb{E}_{\rvz^*}\Big[\mathbb{P}\big(\ry=j \big| \rz^*_{-1,1}=z,\rz^*_{-1,2},...~,\rz^*_{-5,K}  \big) \Big] 
\end{align}

Here we work with the $ \rz^*_{-1,1} $ feature for pure practicality, but the definition is valid for any of the features. The empirical version of the partial dependence function for the same feature is given by the following:
\begin{align}
    \hat{g}_{j,\rz^*_{-1,1}}(z)=\frac{1}{m}\sum_{i=1}^{m} \hat{\mathbb{P}}\big(\ry_i=j ~\big|~ z,z^*_{i,-1,2},...~,z^*_{i,-5,K}  \big)
\end{align}

In this paper, we calculate this function according to the test set data and center each of its summing terms on zero, so it is easier to make comparisons between plots. Thus, we will be interested in average variations in the predicted probabilities of the $ j $ class given variations in an specific feature.

\subsection{Interpretability results}

\subsubsection{What are the filters looking for?}

To better understand the patterns extracted by the neural network's convolutional layer, let us look at the embedding representations of tokens in our vocabulary that have the closest representations to the filters according to cosine similarity. We have nine filters in our model\footnote{That number was chosen during the validation procedure.}, which is a large quantity. For now, we focus on three specific filters (6, 7, and 8), which bring exciting results - the full results will be available in the appendix. Table \ref{tab:interp} shows which tokens\footnote{In the table, tokens were translated from Portuguese to English.} most closely resemble our filters after they are learned.
\begin{table}[h] 
 \centering 
 \caption{Similarity between filters and their most similar tokens. It is possible to check what kind of information the filters seek in a text excerpt by looking at their most similar tokens.} 
 \begin{threeparttable} 
 \medskip
 \begin{tabular}{c|c|c} 
 \toprule 
 \multicolumn{1}{c}{Filters} & \multicolumn{1}{c}{Tokens} & \multicolumn{1}{c}{cos($\theta$)}\\ 
\midrule
\multicolumn{1}{c|}{\multirow{3}[2]{*}{6}} &\textit{"final storage of docket"} &0.46 \\
\multicolumn{1}{c|}{} &\textit{"final remittance to origin"} &0.45 \\
\multicolumn{1}{c|}{} &\textit{"remittance to origin"} &0.42 \\
\midrule
\multicolumn{1}{c|}{\multirow{3}[2]{*}{7}} &\textit{"final storage of docket"} &0.47 \\
\multicolumn{1}{c|}{} &\textit{"temporarily stored docket"} &0.43 \\
\multicolumn{1}{c|}{} &\textit{"final remittance to origin"} &0.42 \\
\midrule
\multicolumn{1}{c|}{\multirow{3}[2]{*}{8}} &\textit{"incident"} &0.55 \\
\multicolumn{1}{c|}{} &\textit{"collect"} &0.5 \\
\multicolumn{1}{c|}{} &\textit{"paycheck"} &0.45 \\
 \bottomrule 
 \end{tabular} 
 \end{threeparttable} 
 \label{tab:interp}
 \end{table}

\begin{figure*}[ht]
   \centering
   \includegraphics[width=.8\textwidth]{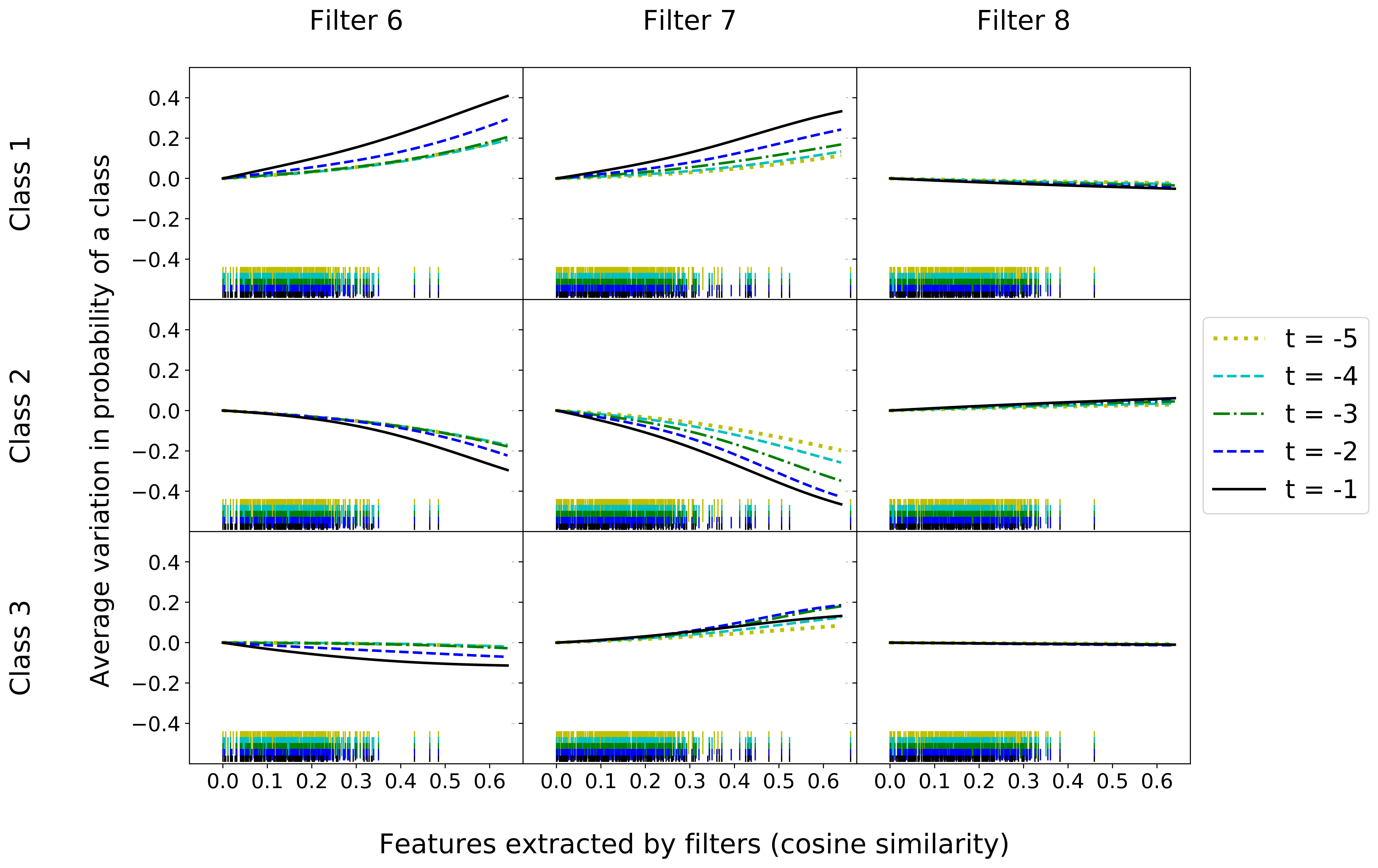}
   \caption{Partial dependence plots: varying features extracted by filters 6, 7 and 8. This plot should be visualized in conjunction with the Table \ref{tab:interp} as you can relate the patterns sought by the convolutional filters, the features extracted by them (cosine similarity), and what is the impact of finding those kinds of patterns with the model's predictions. For example, it is possible to infer from the plots that filter 6 tries to find expressions that distinguish class 1 from class 2 and 3.}
   \label{fig:pdp}
\end{figure*}

One can see that the patterns sought by the neural network do have to do with the classifications we want to make, especially when looking at filters 6 and 7. For example, the expressions "final storage of docket" and "final remittance to origin" indicate archiving of proceedings (class 1), and the expression "temporarily stored docket" may indicate suspension (class 3). We chose to present results for filter 8 because it does not seem to look up important patterns. 
\medskip

\subsubsection{How do features extracted by each filter relate to classification?}

The patterns extracted by filter 6, in Figure \ref{fig:pdp}, explain which legal proceedings are likely to be archived but not suspended or active, which can easily make sense when one sees those expressions linked to filter 6, e.g., "final storage of docket" and "final remittance to the origin." Regarding filter 7, it is possible to notice that the partial dependence functions increase in all plots but the one related to the active proceedings. That fact is understandable because the expressions linked to filter 7, as seen in Table \ref{tab:interp}, are more common to appear when a proceeding is archived or suspended, e.g., "final storage of docket" and "temporarily stored docket." On the other hand, patterns extracted by filter 8, presented in Figure \ref{fig:pdp}, have almost no impact on the neural network's decision as expected.

To conclude this section, we would like to highlight two points that we find most interesting regarding these results. Firstly, the results are intuitive regarding the link between patterns search by the network in the texts and the classification model's output. Secondly, it is possible to notice that more recent information tends to have greater importance in the neural network's decision, what makes sense in the legal context.

\subsection{A note on reproducibility}

In the ideal world, we would like that every time we train our convolutional filters, they return the same final configuration. Unfortunately, this is not possible when the algorithm's seeds are not all fixed, and indeed, this can be a limitation. The good news are that, at least in our experiments, the relevant patterns sought by the filters are pretty consistent; that is, in every experiment we ran, the majority of the relevant patterns are very similar. In a small fraction of the cases, some unexpected relevant patterns show up in addition to the expected ones. This can be due to hidden relations in the data or just due to randomness.

\section{Conclusion}

This work aimed to develop models for the classification of legal proceedings composed of sequential texts in three classes of status (i) archived proceedings, (ii) active proceedings, and (iii) suspended proceedings. Our best performing model achieved accuracy of $.93$ and average F1 Scores of $.89$ (macro) and $.93$ (weighted). The resolution of this problem can help public and private institutions manage large portfolios of legal proceedings in Brazil and possibly other countries, providing gains in scale and efficiency. This paper can also inspire future work involving Law, NLP, and machine learning. Finally, we could extract and interpret the patterns learned by one of our models besides quantifying how those patterns relate to the classification task. The results obtained were satisfactory both in terms of classification and interpretability.

\section{Code and datasets}\label{sec:infos}

The code used in this work as well as the datasets/models can be found partly in \url{https://github.com/felipemaiapolo/predicting_legal_status/} (only small files, including code) or fully in \url{https://bit.ly/3cAl7pD}. The data can also be found in \url{https://doi.org/10.6084/m9.figshare.11750061.v1}.

\section{Computing infrastructure}

AWS VM instance (g4dn.2xlarge) with 8 vCPUs and GPU NVIDIA Tesla T4. GPU was only necessary when making use of BERT.

\section{Acknowledgments}

We would like to thank \textit{Ana Carolina Domingues Borges}, \textit{Andrews Adriani Angeli} and \textit{Nathália Caroline Juarez Delgado} from Tikal Tech for helping us to obtain the datasets. This work would not be possible without their efforts. 

We gratefully acknowledge financial support from Conselho Nacional de Desenvolvimento Científico
and Tecnológico (CNPq), Brazil. Felipe Maia Polo was supported by CNPq during his master's degree while writing this work.

%%
%% The next two lines define the bibliography style to be used, and
%% the bibliography file.
\bibliographystyle{ACM-Reference-Format}
\bibliography{bibliography}

%%
%% If your work has an appendix, this is the place to put it.

\appendix
\section{Appendix}

\subsection{Interpretability}
In this part, we will present the remaining results for the interpretability part. Although there is too much information in this section, we could present the most interesting patterns and results in the paper body.

\begin{table}[H] 
 \centering 
 \begin{threeparttable} 
 \caption{Similarity between most similar tokens and filters 1, 2, and 3} 
 \begin{tabular}{c|c|c} 
 \toprule 
 \multicolumn{1}{c|}{Filters} & \multicolumn{1}{c|}{Tokens} & \multicolumn{1}{c}{cos($\theta$)}\\ 
\midrule
\multicolumn{1}{c|}{\multirow{3}[2]{*}{Filter 1}} &\textit{"halted"} &0.41 \\
\multicolumn{1}{c|}{} &\textit{name of clerk} &0.4 \\
\multicolumn{1}{c|}{} &\textit{"file is sent to attorney's office"} &0.4 \\
\midrule
\multicolumn{1}{c|}{\multirow{3}[2]{*}{Filter 2}} &\textit{"final storage of docket"} &0.47 \\
\multicolumn{1}{c|}{} &\textit{"final remittance to origin"} &0.41 \\
\multicolumn{1}{c|}{} &\textit{"a certain group of companies"} &0.4 \\
\midrule
\multicolumn{1}{c|}{\multirow{3}[2]{*}{Filter 3}} &\textit{"sued financial institution"} &0.4 \\
\multicolumn{1}{c|}{} &\textit{"appear on it"} &0.39 \\
\multicolumn{1}{c|}{} &\textit{"plaintiff-enforced"} &0.38 \\
 \bottomrule 
 \end{tabular} 
 \end{threeparttable} 
 \end{table} 

\begin{figure}[H]
   \centering
   \includegraphics[width=.475\textwidth]{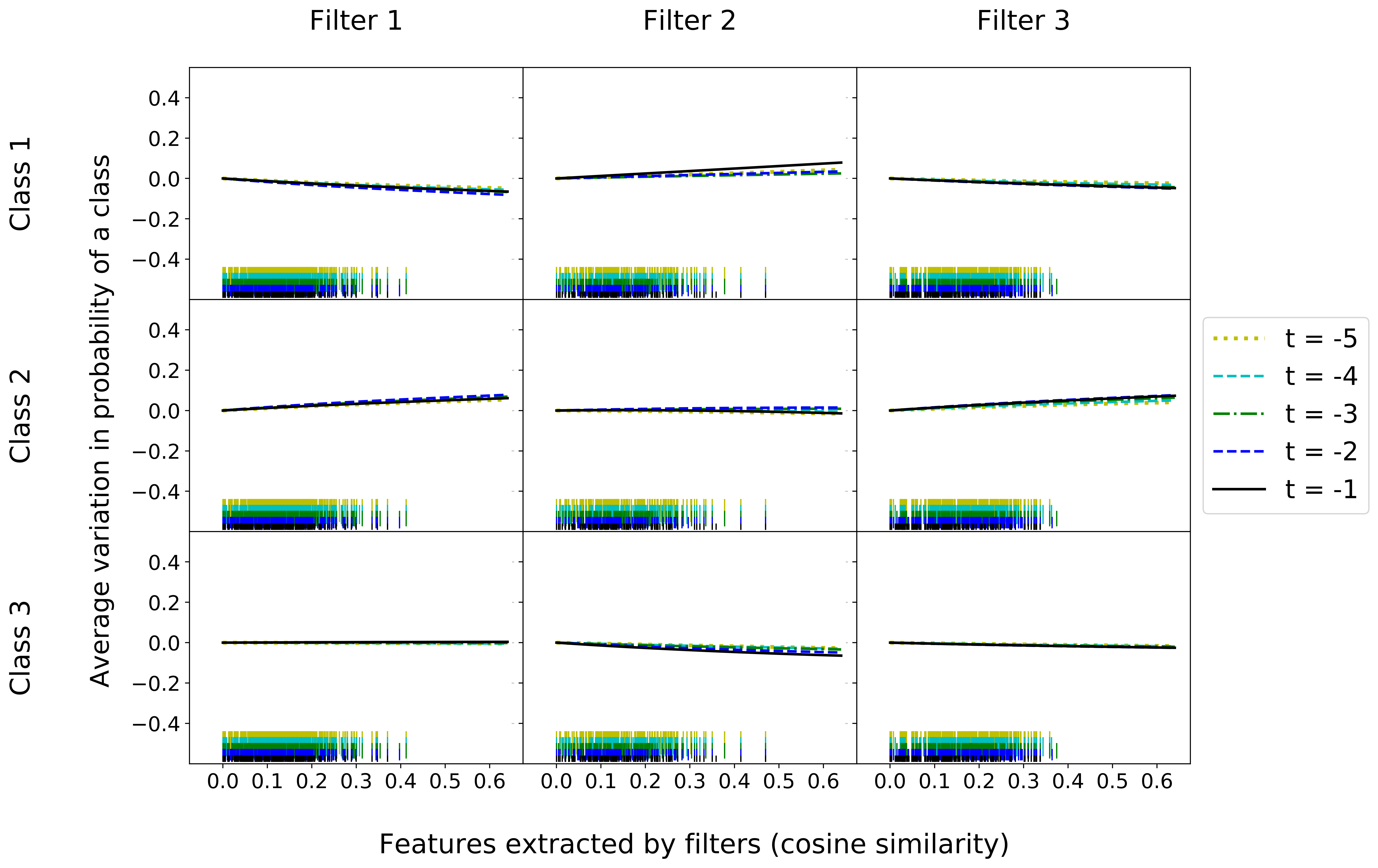}
   \caption{PDP: varying features extracted by filters 1, 2, and 3}
\end{figure}

\begin{table}[H] 
 \centering 
 \begin{threeparttable} 
 \caption{Similarity between most similar tokens and filters 4, 5, and 9} 
 \begin{tabular}{c|c|c} 
 \toprule 
 \multicolumn{1}{c|}{Filters} & \multicolumn{1}{c|}{Tokens} & \multicolumn{1}{c}{cos($\theta$)}\\ 
\midrule
\multicolumn{1}{c|}{\multirow{3}[2]{*}{Filter 4}} &\textit{"of"} &0.42 \\
\multicolumn{1}{c|}{} &\textit{"yours"} &0.41 \\
\multicolumn{1}{c|}{} &\textit{"file sent do judge"} &0.4 \\
\midrule
\multicolumn{1}{c|}{\multirow{3}[2]{*}{Filter 5}} &\textit{"differentiated"} &0.4 \\
\multicolumn{1}{c|}{} &\textit{"fifth panel" (of a Court)} &0.4 \\
\multicolumn{1}{c|}{} &\textit{name of clerk} &0.37 \\
\midrule
\multicolumn{1}{c|}{\multirow{3}[2]{*}{Filter 9}} &\textit{name of clerk} &0.4 \\
\multicolumn{1}{c|}{} &\textit{"will be exempt"} &0.37 \\
\multicolumn{1}{c|}{} &\textit{"automatic manner"} &0.35 \\
 \bottomrule 
 \end{tabular} 
 \end{threeparttable} 
 \end{table}

\begin{figure}[H]
   \centering
   \includegraphics[width=.475\textwidth]{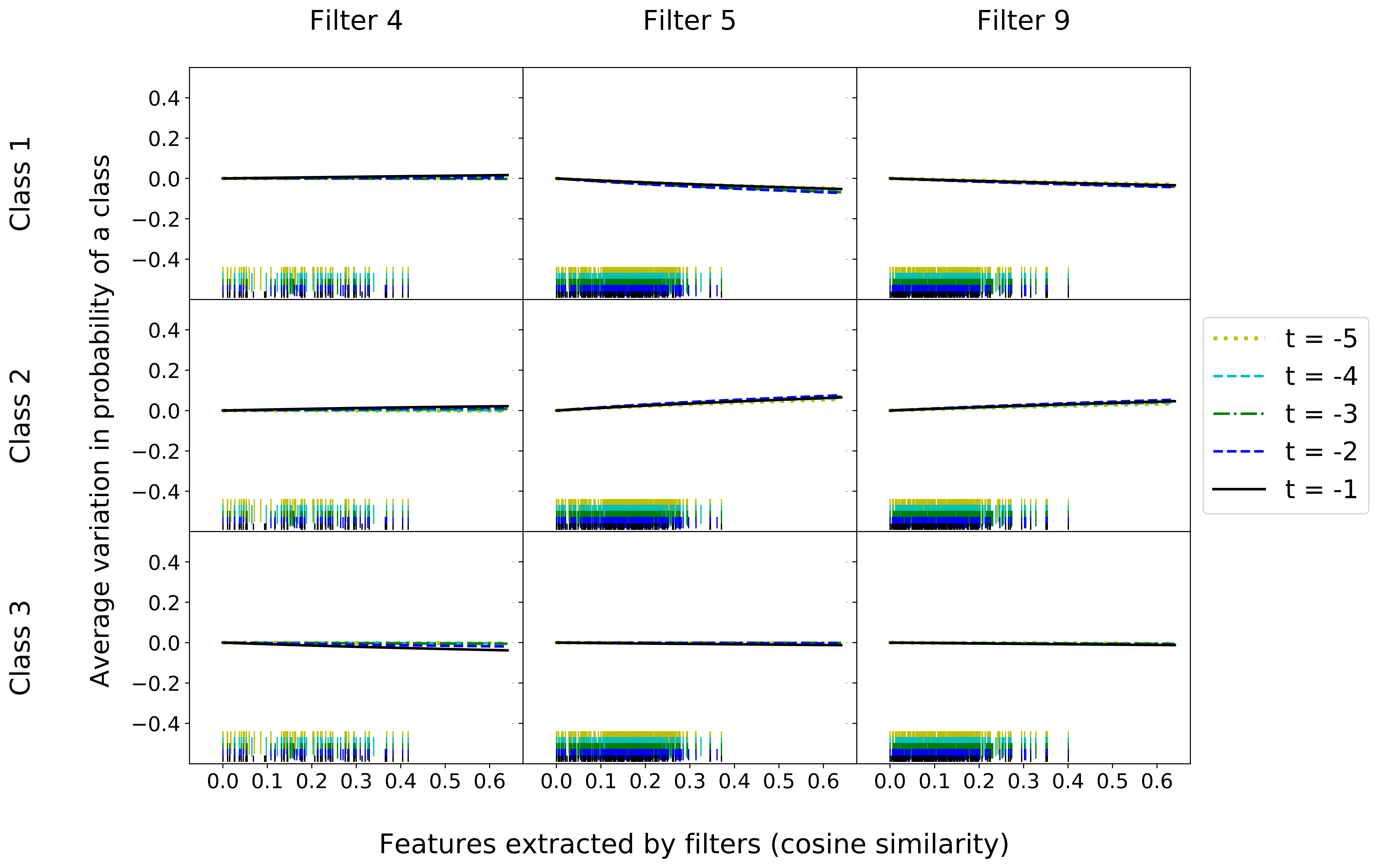}
   \caption{PDP: varying features extracted by filters 4, 5, and 9}
\end{figure}

\subsection{Hyperparameters}

The best values for hyperparameters can be found in Table \ref{tab:hyper}.

\begin{table*}[h] 
 \centering 
 \caption{Best values for hyperparameters} 
  \begin{tabular}{c|c|cccc|cccc|c} 
  \hline 
  Classifier & Feature extraction & Filters & Hidden Size & Reg l1 & Reg l2 & Max. depth & Learning rate & Gamma & Lambda & \\ 
  \hline

\multirow{4}{*}{LSTM}
& W2V & 9 &100 & 0.0005 & 0.0001 & - & - & - & - \\
& Doc2Vec & - &25 & 5e-06 & 1e-05 & - & - & - & - \\
& TFIDF & - &25 & 0.0001 & 1e-06 & - & - & - & - \\
& BERT & - &150 & 0.001 & 0.0005 & - & - & - & - \\
\hline

\multirow{4}{*}{MLP}
& W2V & - &200 & 0.0001 & 1e-05 & - & - & - & - \\
& Doc2Vec & - &200 & 0.0001 & 0.0001 & - & - & - & - \\
& TFIDF & - &200 & 0.0001 & 1e-06 & - & - & - & - \\
& BERT & - &50 & 0.0005 & 1e-05 & - & - & - & - \\
\hline

\multirow{4}{*}{XGboost}
& W2V & - &- & - & - & 5 & 1.0 & 1e-05 & 300.0 \\
& Doc2Vec & - &- & - & - & 5 & 1.0 & 0.01 & 50.0 \\
& TFIDF & - &- & - & - & 7 & 1.0 & 0.01 & 300.0 \\
& BERT & - &- & - & - & 5 & 0.5 & 0.0001 & 10.0 \\
\hline 

 \end{tabular}% 
 \label{tab:hyper}% 
 \end{table*}% 

\end{document}